**North Carolina COVID-19 Agent-Based Model Framework for Hospitalization Forecasting Overview, Design Concepts, and Details Protocol**

Kasey Jones, Emily Hadley, Sandy Preiss, Caroline Kery, Peter Baumgartner, Marie Stoner, Sarah Rhea

RTI International

**Abstract**

This Overview, Design Concepts, and Details Protocol (ODD) provides a detailed description of an agent-based model (ABM) that was developed to simulate hospitalizations during the COVID-19 pandemic. Using the descriptions of submodels, provided parameters, and the links to data sources, modelers will be able to replicate the creation and results of this model.

Keywords: COVID-19, forecasting, hospitalization, agent-based model, ODD

Human subjects' statement: This work was approved by the Institutional Review Boards at the University of North Carolina at Chapel Hill and RTI International.

Funding Sources: North Carolina Department of Health and Human Services Contract #2020-DHHS-COVPS 010; Centers for Disease Control and Prevention Contract #75D30120P07909

Disclaimers: The findings and conclusions in this publication are those of the authors and do not necessarily represent the views of the North Carolina Department of Health and Human Services, Division of Public Health, or the Centers for Disease Control and Prevention.

# 1. Introduction

An Overview, Design concepts, and Details (ODD) protocol provides a standardized way of describing an agent-based model (ABM) so that other scientists can implement the model, promoting open science and replicability.[1,2] This ODD is closely tied to the model source code. The information presented, including methodology, assumptions, and parameters, reflects the state of the model and model source code at the time of this publication. It is subject to change as coronavirus disease 2019 (COVID-19) knowledge evolves and as model updates are made. Although parameters are discussed throughout the ODD, a complete list of parameters and associated references is available in **Appendix A**.

# 2. Purpose and Patterns

The purpose of the model is to simulate demand for hospitalizations and to forecast capacity challenges for North Carolina (NC) hospitals during the pandemic of COVID-19, the disease caused by SARS-CoV-2. The ABM output provides information through near-term (e.g., next 30 days) forecasts of hospital capacities, statewide and by region in NC, to help inform pandemic response efforts. SARS-CoV-2 infection predictions that correspond to different levels of infection spread are used as input for the ABM. The ABM forecasts demand for intensive care unit (ICU) and non-ICU beds for agents with and without COVID-19 (i.e., non-COVID-19 agents hospitalized for other reasons). Stakeholders can use model output to prepare for varying COVID-19 scenarios in NC.

*ABM Patterns*

To validate the hospitalization forecasts, we evaluate the ABM by its ability to reproduce two patterns.

**Pattern 1: Deaths by healthcare facility.** We compare observed data on the typical number of deaths in NC by healthcare facility type to the number of deaths occurring in the model.[3,4]

**Pattern 2: Movement of patients between and to healthcare locations.** We compare observed data on the typical number of patient admissions, transfers, discharges, and steady-state population for each location type to the values produced by the model[5] using patient-level discharge data from seven acute care hospitals of UNC Health from July 1, 2016, to June 30, 2017,[6] and aggregate statewide hospital discharge data.[7,8]

For each pattern, we run the ABM a single time for a 1-year time period. No SARS-CoV-2 infections are introduced during this model run because we compare the ABM's output to data collected pre-COVID-19. If these patterns are accurate, then introducing new agents into hospitals because of the spread of SARS-CoV-2 will allow us to assess capacity at each location.

*COVID-19 Patterns*

The ABM takes as input daily SARS-CoV-2 infection predictions from a set of separate Susceptible-Exposed-Infectious-Recovered (SEIR) models. These predictions are at the NC county level and are for the next 30 days. We assess the ability of the SEIR models to project SARS-CoV-2 infections by comparing the expected growth rate of new infections to the actual growth rate created by the SEIR models after 30 days. The ABM also has specific COVID-19 patterns that are evaluated regardless of the SARS-CoV-2 infection input projections, as described below.

**Pattern 3: SARS-CoV-2 infections.** We compare the daily number of reported COVID-19 cases by NC county from the input data to the number of SARS-CoV-2 infections produced in the model upon initialization and throughout the 30-day model run.

**Pattern 4: COVID-19 hospitalizations.** We compare the proportion of COVID-19 hospitalizations for reported COVID-19 cases in NC with the proportion of COVID-19 hospitalizations to SARS-CoV-2 infections in the model. For this comparison, we include

COVID-19 hospitalizations by agents who were tested and, therefore, recognized SARS-CoV-2-infected agents and those who were not tested and, therefore, unrecognized SARS-CoV-2-infected agents.

## 3. Entities, state variables, and scales

The ABM has two types of entities—agents and locations. Agents represent NC residents and locations include over 500 nodes that are part of a geospatially explicit network and represent healthcare facilities, where agents will seek healthcare during a model run, and the community (Table 1, Figure 1). Agents moving between these nodes represent North Carolinians moving among healthcare facilities and the community. There are four location types: short-term acute care hospitals (STACHs), long-term acute care hospitals (LTACHs), nursing homes (NHs), and the community.

*Table 1. Location nodes of the ABM*

| Location Type | Subtype | Description | Node Count |
|---|---|---|---|
| STACH | UNC Health (UNC)$^{a,b}$ | UNC STACHs[7] | 10 |
| STACH | non-UNC Large$^{a,7}$ | Non-UNC hospitals with ≥ 400 beds[3] | 14 |
| STACH | non-UNC Small$^{a,7}$ | Non-UNC hospitals with <400 beds[3] | 80 |
| LTACH | Not applicable | Long-term acute care hospitals[3] | 10 |
| NH | Not applicable | Nursing homes[9] | 421 |
| Community | Not applicable | Everything outside of the other location nodes | 1 |

$^{a}$The ABM, originally developed to explore healthcare-associated infection prevention interventions in NC, began as a regional healthcare system network model, including 10 modeled UNC Health STACHS.[6] Subsequently, the model was expanded to a statewide platform to include all licensed STACHs in NC (UNC STACHs and all other non-UNC STACHS) in addition to other licensed healthcare facilities (LTACHS, NHs). Here, we maintain the nomenclature developed for the original ABM (i.e., UNC STACHs and non-UNC STACHs).
$^{b}$ Based on available patient-level discharge data from seven acute care hospitals of UNC Health during July 1, 2016-June 30, 2017.[6]

We do not model agent movement among households. Agents located in the community node can be conceptualized to be anywhere in the community other than an STACH, LTACH, or NH.

Both agents and locations are essential in calculating the capacity of healthcare facilities across time.

Healthcare facility locations (i.e., all non-community nodes) have several state variables, including name, physical location (i.e., county), bed count, and a unique location identifier (Facility ID) that is used throughout the ABM (Table 2). Agents who move to one of these locations are given a specific bed within the healthcare facility. At that time, an agent becomes an attribute of the location and if the location is a hospital, is assigned a specific bed. STACH locations have beds designated as ICU or non-ICU and whether a ventilator for that bed is available. Not all NC counties have a modeled healthcare facility. Note that state variables like bed counts, names, and location IDs are not relevant to the community location, and the county attribute is only available for STACHs.

*Table 2. Location state variables of the ABM*

| Location state | Description | Dynamic | Type | Range |
|---|---|---|---|---|
| Bed Count | Number of non-ICU, ICU, and ventilator beds | No | Integer | 1-1000+ |
| Name | Healthcare facility name | No | String | N/A |
| Category | Healthcare facility type | No | String | "UNC", "SMALL", "LARGE," "NH," "LT |
| County | NC county of the healthcare facility | No | String | 1-100 |
| Agents | Current agent IDs and bed assignments | Yes | List | N/A |
| Facility ID | Integer representing the location | No | Integer | 0-536 |

Three of the agent state variables—location, life, and COVID-19 status—are updated daily and used to track how the agents change over time (Table 3). The location state corresponds to the current location for that agent; the life state is a binary variable (i.e., living or dead); and the COVID-19 state represents an agent's current COVID-19 status (susceptible; infected

[asymptomatic or mild to moderate symptoms, severe symptoms, critical symptoms]; recovered). If an agent is assigned a SARS-CoV-2 infection, they will also be assigned a status of tested or untested to reflect if they have a detected or undetected infection, respectively. Note that we assume 100% SARS-CoV-2 test accuracy, although this parameter could be updated in the future. Agents also have several demographic attributes. Throughout the model run, agents who transition locations receive additional state variables. An agent's length of stay (LOS), leave healthcare facility day (model date on which the agent's LOS ends), previous location, and readmission date and location (for agents selected for readmission) may be added, updated, or removed throughout the model run. Note that simulated agent readmission has been turned off to help ease difficulties in calibrating hospitalizations. During initialization, agents are also assigned to presence or absence of comorbidities (labeled as "concurrent conditions" in the model) (see Section: Initialization).

*Table 3. Agent state variables of the ABM*

| Agent state | Description | Dynamic | Type | Range |
|---|---|---|---|---|
| Unique ID | ID for the agent | No | Integer | 1-10,497,471 |
| Age | Age of the agent | No | Integer | 0-93 |
| County | Home County | No | Integer | 1-100 |
| Concurrent conditions (i.e., comorbidities) | Binary variable for presence of comorbidities[10] | No | Integer | 0, 1 |
| Location | Integer of current location | Yes | Integer | 0-536 |
| Life | Life status | Yes | Integer | 0, 1 |
| COVID-19 status | Current COVID-19 status of agent | Yes | Integer | 0-5 |
| ICU status | ICU status | Yes | Integer | 0, 1 |
| COVID test | NA, tested or untested SARS-CoV-2 infection | Yes | Integer | 0, 1, 2 |

*Temporal and Spatial Resolution and Scales*

The ABM is implemented with a 1-day time step, and there is no sense of daily time in the model. After model initialization, the ABM produces 30-day hospital demand forecasts.

## 4. Process Overview and Scheduling

Each time step in the model consists of two substeps, a life substep and a location substep. A life substep represents events causing death for agents and could result in an additional facility bed becoming available (e.g., agent death followed by a discharge from an STACH). A location substep represents events that could result in agents seeking healthcare or changing healthcare facilities (e.g., becoming infected with SARs-CoV-2, changing facility locations, or seeking care for the first time at a healthcare facility). All functions executed in the model for these substeps will randomize the agent order before executing the substep.

The order of updates within the life and location substeps was determined to facilitate the assessment of STACH capacity and, if capacity is exceeded, to count the number of agents that required hospitalization but were turned away from each STACH because no beds were available. The order in which updates (i.e., model functions) are completed determines when a bed becomes available or is filled. Life substep updates are performed first, so that an agent that dies in a healthcare facility will subsequently provide an open bed for another agent.

The location substep follows and occurs in several stages. Agents recovering from COVID-19 or those whose LOS ends for the current day are modeled first. This will open additional beds for future agents. Once all agents who leave a healthcare facility during a day are removed, we model agents who are seeking healthcare facilities. We assumed that healthcare facilities would not transfer an agent if a bed were not available at the next location. Therefore, agents being transferred from one healthcare facility to another occurs next, to prevent any issues with transfers to full hospitals. We then update agents' COVID-19 status after that. We assume that COVID-19 agents would have priority for STACH beds, compared to non-COVID-19 agents; therefore, admission from the community is performed last.

A more detailed description of what occurs within each substep is as follows:

Life substep:

- Simulate agent death. If an agent dies:
  - Record a life state change.
  - For agents in a non-community location, send the agent to the community to make the bed that they were occupying available.
  - Add the agent to the list of agents to recreate (see Section: Submodels).

Location substep:

- Check which COVID-19 agents are set to recover. If they recover, perform a COVID-19 state change.
- Administer a location update (see Section: Submodels) for any agent with a LOS ending on the current day.
- Complete a COVID-19 update (outlined below).
- Administer a location update for any agent selected to leave the community.
- Administer a location update for any agent whose readmission date is for the current day.

To complete a COVID-19 update:

- Using the SEIR infection projections by county, select susceptible, community agents by county to be newly infected.
- Estimate the probability of being tested among newly infected agents. Note that this step does not happen on the first day (see Section: Initialization).
- Assign symptom severity (i.e., asymptomatic or mild to moderate, severe, critical) for newly infected agents.
- Determine hospitalization among newly infected agents.

- Hospitalize COVID-19 agents and assign a LOS.

## 5. Design Concepts

Of all the design concepts available in the ODD protocol, we describe below those that are relevant to the current model.

*Basic Principles*

To be included in future updates.

*Emergence*

To be included in future updates.

*Adaptation*

No adaptive decisions are made at this time.

*Objectives*

Not applicable at this time.

*Learning*

Agents do not learn or adapt over time.

*Prediction*

The ABM does not have explicit or implicit prediction components at this time.

*Sensing*

Agents do not know anything and, therefore, do not use this information.

*Interaction*

Agents in the ABM only have mediated interactions. When an agent occupies a bed in a healthcare facility, other agents do not have access to that resource. This might cause agents to be turned away from an STACH. Because the number of SARS-CoV-2 infections is predetermined using a set of SEIR models, SARS-CoV-2 is not transmitted by agent interaction in the ABM.

*Stochasticity*

To simulate random events that happen to each agent (i.e., death), the ABM compares the probability of events to values generated by a random number generator. If randomly selected, an agent may move location, develop an infection, or die. Random selection of STACH LOS and type of bed assigned (i.e., non-ICU or ICU) also impact stochasticity in the ABM. By setting a random seed, results can be reproduced. All agents that are selected for a specific function, such as agents going to facilities from the community, are randomized before an update takes place.

*Observations*

To be included in a future update.

## 6. Initialization

For each model run, initialization of the agents is the same. In contrast, the projected SARS-CoV-2 infections by county can change across model runs. Parameters that affect an agent's starting location, LOS, and other location or life-specific probabilities do not vary across model runs. The only difference across model runs is what happens after the model is initialized, allowing stakeholders to focus on model output related to healthcare demand and STACH capacity given changes to SARS-CoV-2 infection counts resulting from proposed interventions. Location entities are created using input data that provide healthcare facility names, bed counts, and other attributes. Before agents are initialized, the location entities do not have any agents (i.e., all beds are empty).

Agents for the ABM are created using demographic variables provided by the RTI SynthPop™.[11] This file consists of one anonymized synthetic person per row, containing the synthetic person's home county; Federal Information Processing Standard code (corresponding to one of NC's 100 counties); sex (female, male); and age in years. We completed two preprocessing steps on this file to prepare it for model input:

1. The 2017 NC synthetic population for 2017 has approximately 9.3 million agents. We randomly selected approximately 1.2 million agents and duplicated them to create approximately 10.5 million agents for the simulation, reflective of the current NC population.[12] Only agent demographic information such as age, sex, and general location are used; duplicating agents is acceptable here because no household-specific attributes (e.g., household size) are used.

2. We binned the age of the agents in this expanded file into age groups (<50, 50-64, ≥ 65) to match the age groups used in previous models.[6]

For each row in the augmented version of the input synthetic population file, an agent is initiated with the variables provided in that row. Agents are then randomly assigned a concurrent conditions binary category (i.e., comorbidities) based on their age group.[10] Unless selected to start in a facility, all agents are initially set to start in the community.

All NH and LTACH facilities do not rely on model parameters for their initial capacity levels. At initialization, agents are assigned to these facilities until they reach 70% capacity (expert input). The agents selected to start in these facilities are randomly selected based on a probability distribution formed by taking the relative distance between the facility and each county. Counties close to the facility will be pulled from more often. For NHs, agents must be ≥ 65 years of age to be selected.

Each STACH in the model is initiated with ICU and a non-ICU agent to match starting capacity percentages specified by either model parameters or by real-time capacity percentages for each hospital. The ABM currently initiates normal STACH beds at 65% capacity and ICU beds at 54% capacity[13] by default. Alternatively, in cases where individual hospital capacity data are available from another source, the model can initialize the non-ICU and ICU capacities to the provided levels. Based on these starting capacities, agents are selected to start in STACHs instead of the community. Agents are selected based on their home NC county of residence. STACH discharge data[7,8] inform the model on which counties can supply agents for each hospital. For initialization, all counties that had previously supplied agents to an STACH are treated equally. This means that all agents from any county listed in the discharge data for a given hospital have an equal chance of being selected. Other than home county, age also affects the agent's probability of being selected for hospitalization. Agents are weighted by their age, such that 40% of hospitalizations are <50, 20% are 50-64, and 40% are ≥ 65 years of age. This distribution of ages matches the distribution found in the aggregate hospital discharge data.[7] All agents who were assigned a starting location other than the community are assigned a LOS. Agents initialized in a facility are assumed to have already completed part of their stay. NH LOS values are drawn from the NH LOS alternative file (see Section: Input Data). STACH and LTACH LOS values are sampled from a small simulation which is run at the beginning of the model run. For each facility, LOS values are sampled from the facility's LOS distribution, and then these "patients" are moved forward in time a set number of days, subtracting one from each value each day and replacing values that reach 0 with new values sampled from the distribution. The length of the simulation is determined by calculating the 95th percentile for the LOS of the initial sample. This simulation produces a list of days that is then used to sample LOS values for the facility on Day 0. Conceptually, this calculates a *remaining stay time* for existing patients, as opposed to a new LOS value.

All agents are initialized as susceptible to SARS-CoV-2. During the COVID-19 step on Day 0, agents equal to the cumulative number of SARS-CoV-2 infections in NC are assigned SARS-CoV-2 (i.e., become a COVID-19 agent). Throughout the model run, agents must be in the community to be assigned SARS-CoV-2 and, subsequently, become a COVID-19 agent. Using the length of exposure and length of infection from the SEIR model, we estimate that a small portion of these cumulative infections will still be currently infected (i.e., "live infections"; see Section: Submodels). Agents that are randomly selected as inactive are then set to recovered, while agents that are still active infections may be randomly assigned hospitalization. On day 0, COVID-19 hospitalizations are forced to match the known hospitalizations at the date the model was initialized. By default, if no hospital-level information is provided, it is assumed that there are 1,000 COVID-19 hospitalizations distributed evenly across non-ICU and ICU beds in all hospitals in the state. This value was chosen based on average statewide hospitalization counts publicly reported in NC between March and October 2020. If hospital-level COVID-19 hospitalizations are available, then the model uses these counts directly to assign initial COVID-19 hospitalizations in each hospital. In either case, the model takes the number of expected non-ICU and ICU COVID-19 hospitalizations on Day 0 for each hospital and scales them to account for the model population and the number of available beds in the hospital given the non-COVID-19 agents already admitted. As each county is assigned SARS-CoV-2 infections, these infected agents (i.e., COVID-19 agents) have a chance of being considered for filling a COVID-19 hospital bed which is slightly higher than the statewide percent hospitalized (to make sure all needed beds are filled). If there is an unfilled SARS-CoV-2 bed needed for a hospital within 60 miles of an agent selected for hospitalization, they are assigned that bed. Hospitals are considered in the order of increasing distance from the agent's county center using the County to Facility Distances input data file.

## 7. Input Data

Data collected to calibrate previous iterations of the ABM[5,6] are not described in this section. All references to data files can be found on the models repository[32] at the location specified. The repository contains all code and data necessary to reproduce each file.

*Location Transitions (*data/transitions/location_transitions.csv*).*

Once it is determined that an agent is moving to another location, the probability of transferring from an agent's current location to another healthcare facility type is outlined in the location transitions file. Probabilities are based on an agent's home county and age group. Although unique transition probabilities are available at the individual STACH level, other location types (community, NH, and LTACH) have a single row of probabilities in this file. Transition probabilities must sum to 1 and are based on discharge data for each healthcare facility type. For example, the six transition probabilities for a UNC STACH are based on the total discharges that specific STACH had to each of the six location types (community, UNC, Small non-UNC, Large non-UNC, NH, LTACH).[7] Individual facility-level data were not available for NHs and LTACHs. Therefore, we estimated the total number of transitions between healthcare facility types by assuming that the total discharges to NHs and LTACHs was equal to their total admissions. See Table 4 for a list of parameters to determine transitions between specific STACH types, which was used to create the location transitions file.

*Table 4. Parameter values for transitions between STACH types*

| Parameter | Proportion | Context |
|---|---|---|
| Large non-UNC to Large non-UNC | 0.8 | Proportion of patients leaving a large STACH that are selected to go to a non-UNC STACH and go to a large non-UNC STACH |
| Large non-UNC to Small non-UNC | 0.2 | Proportion of patients leaving a large STACH that are selected to go to a non-UNC STACH and go to a small non-UNC STACH |
| Small non-UNC to Large non-UNC | 0.9 | Proportion of patients leaving a small STACH that are selected to go to a non-UNC STACH and go to a large non-UNC STACH |
| Small non-UNC to Small non-UNC | 0.1 | Proportion of patients leaving a small STACH that are selected to go to a non-UNC STACH and go to a small non-UNC STACH |
| non-UNC to UNC | 0.0322 | Proportion of non-UNC discharges that go to a UNC STACH |
| UNC to UNC | 0.90 | Proportion of UNC discharges that go to a UNC STACH |

*Community Transitions (*data/transitions/community_transitions.csv*)*

The probability of leaving the community on any given day and being admitted into a healthcare facility is outlined in the community transitions file. These probabilities are based on an agent's home county and age and were determined by assessing how many people were admitted to each facility over a 1-year time frame.[3] Note that only agents ≥ 65 years of age can be admitted to a NH. This file contains a column with the daily probability that an agent (from a specific county and in a specific age group) would leave the community for admission to any healthcare facility. In addition to this daily probability, there are five individual probabilities for being admitted to each (non-community) healthcare facility type. These five probabilities must sum to 1. To calculate these probabilities, we evaluated the total expected admissions to each healthcare facility type from the community for each age and county combination.[3,7–9,13]

*Facility Discharge Data (data/discharges/county_discharges.csv)*

For each STACH type there is a file containing the number of agents that were discharged from each hospital by county.[7] These files are used to create the healthcare facility transition files. Discharge data by county of residence and hospital only include counties that account for ≥ 1% of discharges.

*County-to-Facility Distances (data/geography/county_…_distances_sorted.json)*

For each county and each facility type (STACHs, NHs, large non-UNC STACH, and LTACHs), there exists a dictionary of distances between that county and all facilities. Distance was calculated using the centroid of the county and the facility addresses.

*NH LOS (data/cLOS_MDS_2016.csv)*

The NH LOS file contains the number of NH patients that were discharged from NHs for each possible length of stay between 0 and 2,000+.[14] These data are converted into a list of potential LOS values and are used when assigning the LOS for a newly admitted NH agent.

*NH LOS: Alternative (data/nh_time_until_leaving.csv)*

The NH LOS alternative file contains a list of how many days remain in agents' NH LOS values at the end of a test model run. This file was created by running a test model for 2 years and taking a snapshot of how many days remained in each NH agent's LOS. Because this file is used to initialize NH LOS in the ABM and was not available when the test model run was completed, NH agents' initial LOS values were assigned using the normal NH LOS file.

*Facility Files (data/locations/…)*

There are several files containing facility information used as input into the ABM. These files are crosswalks between the Name of the facility and the actual facility ID. The hospital facility file also contains the number of non-ICU and ICU beds for STACHs.

*COVID-19 Reported Cases (SEIR/)*

This file contains the number of confirmed COVID-19 cases by county and by day. The most recent data available should be used. Instructions for downloading and cleaning these data are found in the SEIR directory of the public repo.[32]

*Hospital Census Data*

This file contains the number of occupied inpatient beds and ICU beds and counts of confirmed and probable inpatient and ICU COVID-19 patients by hospital. As access to these data is limited; alternative expert-informed defaults are included in the public repository version of the ABM.

## 8. Submodels

The ABM has three submodels, one for each state variable: life, location, and COVID-19. We also outline the SEIR model that is run before the ABM begins in this section. If another outside model is used for creating SARS-CoV-2 projections, the final subsection of Section: Submodels can be skipped.

*Life*

Death is an important component of the ABM. The model is parameterized to certain hospital capacity levels, and without death, these levels would not be maintained. Hospital beds would also not become available as often as they do in reality if we did not model death.

The daily probability of death occurring from natural causes was derived using CDC WONDER data[4] and is included in the input parameters file. Death probabilities are provided by age group. To calibrate to the number of expected deaths in each facility type,[3] we included multipliers in the parameters file for each facility type. These multipliers can be used to increase or decrease the frequency of deaths in the model and were raised and lowered during calibration until the correct number of deaths occurred for each facility type.

We do not specifically model death from COVID-19 within the ABM at this time, although this is an addition we plan to make in the future.

*Recreate Agents*

Every 15 days in the model, the attributes of deceased agents are used to create new agents. Recreating agents every 15 days allows the model to skip over this function until enough agents have died that we need to recreate them. All regenerated agents start in the community, do not have COVID-19, and have no additional location attributes (e.g., current LOS, readmission date). These agents are appended to the list of agents in the model and are assigned a unique ID based on the number of agents in the model. This process is only used to maintain the NC population throughout a model run and was primarily used during model calibration. As the COVID-19 runs only last 30 days, this step has little to no impact on the capacity estimates.

*Location*

The location submodel controls agent movement. Agents move locations if one of five situations occurs: (1) Agents in the community are randomly selected to move based on their daily probability of movement; (2) Agents who previously left a healthcare facility, moved to the community and were given a readmission date return to that healthcare facility on their readmission day; (3) Agents in healthcare facility nodes move if their LOS ends; (4) Agents in healthcare facility nodes who die are moved to the community; or (5) Agents who are assigned COVID-19 are randomly assigned to seek hospitalization.

Each agent in the community has a daily probability of moving from the community. In the community movement step, the probability (assigned using the community transitions file) for each agent is compared to a random number to see if that agent will leave the community. If selected, a second random number is drawn to determine which type of healthcare facility the agent will move to (based on the location transitions file). Finally, based on the type of healthcare facility selected, the agent is assigned a specific healthcare facility ID based on their home county (using the healthcare facility transitions file).

Agents in healthcare facilities only leave that facility if their LOS ends or if they die. For agents whose LOS ends, we use their location transition probabilities to determine the healthcare facility type of their next destination. A random probability is generated, and this probability is compared to their transition probabilities. Most agents will move to the community upon discharge, but some are selected for transfer. Once a healthcare facility type is determined, if a non-community node is selected, we compare a second random number to the facility transitions to determine the exact facility ID. There is one hard-coded component that is added to this logic for NH agents. For agents who previously transitioned from a NH to an STACH, we assume ~80% will return to the previous NH when their STACH LOS ends.[15] We included this hard-coded component in the face of a general lack of available data on NH agent movement; this can be updated in the future with additional data. When, because of censored values, the aggregate hospital discharge data cannot be used, we apply different restrictions for movement according to the healthcare facility type, as described in more detail below.

- **Small (<400 beds) non-UNC STACHs:** Small non-UNC STACH discharge data are available for 99 NC counties. The ABM uses distributions created from these available discharge data[7] to randomly assign agents selected to move to a small non-UNC STACH. Agents that are discharged from a small non-UNC STACH can be selected to move to another small non-UNC STACH, with assignment based on the distributions. Agents are not permitted to remain at the same small non-UNC STACH once their LOS is complete. Rather, in this rare situation, if an agent is selected to transfer from a small non-UNC STACH to another small non-UNC STACH and that agent is from a county with discharge data available for only one small non-UNC STACH, the ABM randomly assigns the agent to another small non-UNC STACH.

- **Large (≥ 400 beds) non-UNC STACHs:** Large non-UNC STACH discharge data are available for 74 NC counties. The ABM uses distributions created from these available discharge data[7] to randomly assign agents selected to move to a large non-UNC STACH. If an agent who is selected to move to a large non-UNC STACH is from a county with no large non-UNC STACH discharge data available, the ABM assigns the agent to a large non-UNC STACH based on a probability distribution that weights the size of available facilities with the relative distance to the agent's county. The relative importance of bed count versus distance are parameters given to the model. Similarly, if an agent is selected to move from a large non-UNC STACH to another large non-UNC STACH, the ABM uses the distributions, followed by the probabilistic approach, to assign the agent. In the rare situation that both these methods fail, the ABM randomly assigns the agent to a large non-UNC STACH.
- **UNC STACHs:** Agent movement to UNC STACHs is based on the patient-level discharge data from seven UNC Health acute care hospitals during June 30, 2016-July1, 2017, for each of the 10 modeled UNC STACHs which serve a 41-county catchment area.[6] Most of the agent movement to and from UNC STACHs is completed by agents whose home county is among the 41-county catchment area. The ABM uses distributions that we created from the available discharge data to select an agent's initial UNC STACH and inform its movement from one UNC STACH to the next (i.e., transfer). If an agent is selected to transfer from a UNC STACH to another UNC STACH and that agent is from a county with discharge data available for only one UNC STACH, the ABM randomly assigns the agent to one of the two largest UNC STACHs.

When an agent arrives at an STACH or LTACH, the agent is assigned a LOS based on a gamma distribution unique to the healthcare facility. We use patient-level data, available for 7 of the 10 UNC STACHs, to obtain STACH-specific LOS gamma distributions. For the remaining three UNC STACHs and the non-UNC STACHs, for which patient-level data were not available, we used aggregate discharge data to estimate the parameters of a gamma distribution.[8] When an agent arrives at a NH, the agent is assigned a LOS based on a list of possible LOS values for NH agents (see Section: Input Data).

Non-COVID-19-related readmission is unique to STACHs, and several agents will be randomly selected for readmission at the time of leaving an STACH. Thirty-day readmission is approximately 10% and is based on the patient-level UNC Health data. If selected for readmission, the agent is randomly assigned a readmission date between 1 and 30 days from the current day. It is possible that an agent will be readmitted during this 30-day window. If this occurs, their previously assigned readmission date is deleted.

The final way that an agent can move to a healthcare facility is when they are assigned COVID-19. Hospitalization rates for COVID-19 agents are based on their COVID-19 state (asymptomatic/mild to moderate, severe, critical) and a Bayesian approach to incorporate agent age and comorbidities to inform these estimates, described in detail in the COVID-19 submodel section below. All COVID-19 agents with severe or critical symptoms and if the parameters allow, a random selection of agents categorized as asymptomatic or mild to symptoms are hospitalized and assigned an LOS (see **Appendix A**). Their hospital is assigned using the same methods described for non-COVID-19 agents seeking hospitalization from the community.

Agents who arrive at hospitals for admission can be turned away if the STACH is at capacity (i.e., all beds are occupied by other agents). When determining that an agent is moving to an STACH, the ABM also assigns an ICU flag and a ventilator flag to indicate which type of STACH bed is needed. Non-COVID-19 agents have ICU status determined by a logistic regression model that uses their age, presence/absence of comorbidities, and their assigned LOS to determine a probability of receiving ICU care (derived from patient-level UNC Health data). ICU beds and ventilator assignments for COVID-19 agents are based on the agent's COVID-19 disease category, as described in the COVID-19 submodel section. If an STACH does not have a bed that matches an agent's need, the agent will attempt up to three additional methods for finding an STACH to meet that need, as follows:

1. The agent will try their second choice STACH, based on the STACH probabilities for that agent.
2. The agent will try any other STACH with a catchment area that includes the agent's home county.[7]
3. The agent will try any additional NC STACHs located within a 200-mile radius of the centroid of the agent's home county.

If the agent is turned away from their first choice STACH, that agent is added to a list of agents turned away. In this list the model maintains the date, location, and county of agents that were turned away during a model run. If the agent is turned away from all STACHs that they tried, that agent is added to a list of agents who were completely turned away during the model run. Agents who are randomly selected to transfer from another healthcare facility will only try their first choice STACH. If this STACH is at capacity, the agent returns to the community. We based this assumption on the premise that a healthcare facility would not transfer an agent to an STACH that did not have a bed available.

*COVID-19*

The COVID-19 submodel is a smaller model within the larger ABM that simulates COVID-19 and monitors COVID-19 status among agents. This submodel is primarily used to estimate demand for non-ICU and ICU hospitalization among all agents, including COVID-19 agents. The submodel also includes preliminary estimates for ventilators and demand for NH beds following hospitalization with COVID-19. In the future, these estimates could be extended to include items related to hospitalization, including demand for personal protective equipment and healthcare staffing needs.

The agents in the COVID-19 submodel are updated daily. Agents transition between COVID-19 disease states (i.e., susceptible, infected, recovered) and hospitalization.

**Summary of 30-Day COVID-19 Submodel Run**

The agents in the COVID-19 submodel are updated daily. Agents transition between COVID-19 disease states (i.e., susceptible, infectious, recovered) and hospitalization.

*Day 0*

Day 0 is unique compared to the other days of the model because it is equivalent to the most recent date of the available NC county-level COVID-19 reported case data used in the model. On Day 0, the model accounts for all agents who are currently susceptible, infected, or recovered and for the number of COVID-19 agents who are currently hospitalized. To determine the number of agents that have been or are currently infected (i.e., cumulative infections), the model evaluates the cumulative number of COVID-19 reported cases in NC since the date of the first reported case in NC and applies a multiplier to reflect the likely number of unreported infections.[16–19] The sum of the reported cases and estimated unreported infections is the count of cumulative infections.

The SEIR model (described below in **SEIR**) is used to separate the cumulative count of infections into the count of estimated current, or "active," infections and the count of estimated recoveries. These counts are used to select the agents currently infected on model-day Day 0 and the agents currently recovered on model-day Day 0. The process used to select agents for infection is described below in **Select Infected Agents**.

On Day 0, the number of hospitalized COVID-19 agents is calibrated to the census of probable and confirmed hospitalized COVID-19 patients in the hospital census data. These estimates are at the hospital level. In the absence of hospital census data, a default of 1,000 hospitalized COVID-19 agents is used. The process described below in **Assign Symptom Severity** is used to determine the probability of hospitalization for each agent with these probabilities scaled on Day 0 to ensure that the number of hospitalized COVID-19 agents on Day 0 approximately matches the actual number of hospitalized COVID-19 patients in NC. At the end of Day 0, the appropriate number of agents have been assigned susceptible, infected (with or without hospitalization), and recovered states to approximately match the estimated current infections, recoveries, and hospitalizations from COVID-19 in NC. We currently do not account for COVID-19-associated death in the model.

*Day 1 through Day 30*

Starting on Day 1, agents move through different COVID-19 disease states and periods of hospitalization. On each day, the following actions are completed:

- Check for Recovery from COVID-19 Among Previously Infected Agents
- Select Agents to be Newly Infected

- Estimate Likelihood of Testing Among Newly Infected Agents
- Assign Symptom Severity (asymptomatic or mild to moderate, severe, critical) for Newly Infected Agents
- Determine Hospitalization Among COVID-19 Agents
- Hospitalize Agents and Assign LOS

These steps are described in greater detail below.

Check for Recovery from COVID-19 Among Agents Hospitalized on the Previous Day

On Days 1–30, the model first checks whether agents who were previously infected have recovered. This is an important step because recovered agents either leave their STACH beds (if they were admitted) or are no longer in need of an STACH bed (if they were not admitted anywhere because of capacity limits). This step is completed first each day to determine the estimated number of available beds at each STACH each day.

Upon assignment of infection (see **Select Infected Agents**), each agent receives an estimated recovery day. If the agent was admitted to an STACH, this recovery day is equivalent to their LOS (see **Hospitalized Agents and Assign Length of Stay**). If the agent was not admitted to an STACH, this recovery day is equivalent to the length of their infection. On the agent's recovery day, the state of the agent is changed to RECOVERED and the agent is removed from the STACH bed (if applicable) and marked as no longer in need of an STACH bed. On the agent's recovery day, the state of the agent is changed to RECOVERED and the agent is removed from the STACH bed (if applicable) and marked as no longer in need of an STACH bed.

Some proportion of COVID-19 patients require care in a NH following hospital discharge which might be related to the level of care they required during hospitalizations (i.e., ICU-level care vs. non-ICU-level care). For each recovered COVID-19 agent that is discharged from an

STACH, the model uses a probability of needing NH care that is dependent on whether the agent received ICU care while hospitalized (expert input) (**Appendix A**). The model counts the number of recovered COVID-19 agents discharged from an STACH to a NH based on these probabilities (see **Appendix A**). Recovered COVID-19 agents are assumed to remain in a NH for the duration of the model run (default 30 days).

COVID-19 agents with the status RECOVERED are not eligible to be reinfected. This assumption can be updated as more information becomes available regarding reinfection.

**Select Infected Agents**

On each day, the model selects agents to be infected with SARS-CoV-2. This step requires the count of the number of agents to be infected by NC county. Day 0 is unique, as described in **Summary of 30-Day COVID-19 Submodel Run: Day 0**. The count of infected agents on Day 0 is the number of active infections as calculated with the estimated cumulative infections and the SEIR model. From Day 1 to Day 30, the count of new infections by day for each county is estimated using the county-specific SEIR models (described in **SEIR** below).

Only agents that have a status of SUSCEPTIBLE and are located in the community in a given county are eligible for infection. We do not currently model SARS-CoV-2 infections originating in healthcare facilities, although this is an area for future work. In the rare case that, on any given day the SEIR model estimates more infections than there are eligible agents, all remaining eligible agents in a county will be infected. This assumption will be revisited in the future.

The probability of an eligible agent being assigned COVID-19 is also dependent on the age of the agent. These probabilities are derived from observed age data among COVID-19 reported cases in NC and are described in further detail in **Appendix A.** At the end of the Select Infected Agents step, the appropriate number of agents are infected as per the estimates from the SEIR model and the distribution of ages among the newly infected agents is congruent with the reported NC data from positive tests as described in **Appendix A**. Parameters and assuming that the distribution also applies to the agents created using the reported case multiplier.

**Estimate Probability of Testing Among Newly Infected Agents**

On Days 1-30, agents newly infected with SARS-CoV-2 are identified as either *tested and reported* or *untested and unreported*. We model testing status because we assume that *tested and reported* infections are more likely to be hospitalized than *untested and unreported* infections. A basis for this assumption is that infections with more severe symptoms that would lead to hospitalization are more likely to be tested (expert input). Note that an *untested and unreported* infection could become a *tested and reported* infection, particularly if an agent is hospitalized and subsequently tested in the STACH. This change in status is not currently tracked in the model.

The probability of being tested is dependent on the overall observed ratio of tested and reported cases to unreported infections. This parameter is described in further detail in **Appendix A**. The model includes functionality to modify this parameter by age if evidence suggest that the probability of getting tested differs by age group.

Testing status is not used on Day 0 because the number of hospitalizations is forced to match the observed reality on Day 0, rather than based on probability of hospitalization as is done for Days 1-30.

**Assign Symptom Severity**

Every agent with an active infection is also assigned a symptom severity level. Each symptom severity level is associated with a level of hospitalization. These symptom severities are assigned the same day as the infection starts for that agent and persists throughout the duration of infection for that agent until the agent's day of recovery (described in **Check for Recovery from COVID-19 Among Agents Hospitalized on the Previous Day**). Further development will explore opportunities to transition between symptom severity levels.

The levels of symptom severity in the model include the following:

- Asymptomatic and Mild to Moderate: symptoms do not require hospitalization
- Severe: symptoms require hospitalization in a non-ICU bed
- Critical: symptoms require hospitalization in an ICU bed (possibly with a ventilator)

The probability of symptom severity level is impacted by the age of the agent and whether the agent has comorbidities.[10] These probabilities are described in greater detail in **Appendix A**. On Day 0, the distribution of symptom severity levels is scaled to ensure that the number of Severe or Critical COVID-19 agents in the model matches the number of actual COVID-19 patients hospitalized in NC per hospital occupancy data.

**Hospitalize Agents and Assign Length of Stay**

COVID-19 agents seek an STACH based on their symptom severity level. Agents with *Severe* symptoms will only seek a non-ICU bed and will attempt multiple STACHs as described in **Location** until admitted or completely turned away. Agents with *Critical* symptoms will only seek an ICU bed and will attempt admission at multiple STACHs as described in **Location** until admitted or completely turned away. A proportion of agents with *Critical* symptoms are also flagged as in need of a ventilator. This proportion is further described in **Appendix A**.

If a COVID-19 agent is admitted to an STACH, their LOS is assigned based on observed data (as described in **Appendix A from COVID-19 Among Agents Hospitalized on the Previous Day**). If a COVID-19 agent is completely turned away from all attempted STACHs, they are not assigned a LOS but, instead, assigned a COVID-19 recovery day based on the observed duration of the infection (see **Appendix A**).

**Variables Collected from COVID-19 Model**

The variable categories collected for each day of the forecast in each model run are presented in Table 5.

*Table 5. Variable categories collected for each day of the forecast*

| Variable Category | Options | Description |
|---|---|---|
| Infection Counts | New and Cumulative | New and cumulative SARS-CoV-2 infection counts by day of the model |
| Seeking Hospital Bed | All and COVID-19 specific; non-ICU and ICU | Count of agents (both non-COVID-19 and COVID-19, or COVID-19 specifically) seeking an STACH bed |
| Hospital Census | Non-COVID-19 and COVID-19; non-ICU and ICU | Census of currently hospitalized agents, either with or without COVID-19 and either non-ICU or ICU |
| Patients Turned Away | Non-COVID-19 and COVID-19; non-ICU and ICU | Count of COVID-19 agents that were completely turned away from all STACHs where they attempted to be admitted |
| Demand | Total and COVID-19 specifically; non-ICU and ICU | Count of agents (total or COVID-19 specifically) in need of a bed on any given day |

*SEIR*

The ABM relies on daily SARS-CoV-2 infection projections. To run the model, we provide a submodel for creating these case projections. A SEIR model is a deterministic compartmental model used to simulate the spread of infectious disease. Because the ABM is being used to estimate infections and hospitalizations during an ongoing pandemic, we have altered the SEIR approach to account for available past data. Using a single effective reproductive number, we run an individual SEIR model for each of NC's 100 counties (Table 6).

*Table 6. SEIR model parameters*

| Parameter | Value | Context |
|---|---|---|
| Case Multiplier | 10 | Multiplier representing unreported SARS-CoV-2 infections[16–19] |
| Percent Susceptible | 90% | The percent of the population that remains susceptible when the simulation starts; based on 9.4% antibody positivity rate[20] |
| Length of Infection | 6 Days | Average length of infection[21] |
| Length of Exposure | 5 Days | Average number of days after exposure before someone becomes infected[21] |
| Population | Varies | 2017 Population of the county[12] |
| $R_e$ | Varies | Effective reproductive number |

A typical SEIR model starts on Day 1, with only an estimated number of initial infections. To initiate our SEIR model that starts after the pandemic has already been ongoing, we estimate each compartment, S, E, I, and R, based on the number of COVID-19 reported cases from 03/03/2020 through the most recent day. Using the case multiplier parameter, we multiply all reported new cases by this multiplier to get an estimate of the actual number of SARS-CoV-2 infections that have occurred.[16–19] This process is completed for each day, $k$, such that:

- The estimated infections are added to $I_k$.
- $E_{k-i}$ is updated to reflect the number of exposed individuals required to achieve the level of new infections for day $k$.
- $R_k$ is updated based on the amount of people recovered from $I_{k-1}$.
- $S_k$ is equal to 1 minus the sum of the rest of the compartments.

Transition from recovered to susceptible is not modeled currently but can be revisited as more information becomes available about reinfection. If there have been no COVID-19 reported cases for a specific NC county, we set the proportion of infected individuals for the last day of available data to be (1 / county population).

Before the SEIR model begins, using the current day as day 0, we set the susceptible proportion equal to the percent susceptible parameter, taking or adding an equal proportion from the recovered proportion. This allows the SEIR to start with the expected proportion of susceptible agents, as the pandemic has already been ongoing. The estimates for the exposed and infectious compartment remain unaltered. For Day 1 going forward, we use the standard SEIR method to model transitions between each SEIR compartment. SEIR models use a Beta*S*I calculation to determine the proportion of newly exposed agents.[22] The beta parameter is based on the estimated $R_0$ value. For each county-level SEIR model, the input $R_e$ is updated using what we call a county correction.

For a model run, we are estimating the number of new infections for the next 30 days given the specified $R_e$ value. Because each NC county has a different COVID-19 reported case count and will respond to interventions in different ways, we introduced a county correction into the SEIR model. This correction is based on comparing the estimated growth rate for the state over the past 2 weeks, $State - R_e$, to the estimated $R_e$ for each individual county, $R_{e_i}$. Counties that have had higher $R_e$ values are assumed to continue to have higher than average SARS-CoV-2 spread during the next 30 days. To arrive at the final $R_e$ used in the SEIR model for each individual county, we use $CountyR_e = R_e * StateR_e/R_{e_i}$. SEIR models, however, do not use $R_e$ values. To estimate the $R_0$ for a specific county (the $R_0$ is in a totally susceptible population), we take the input $R_e$ and divide it by the remaining proportion of susceptible individuals at time $K$, the last date of available data.

The final estimates of the SEIR model produce the number of estimated new reported COVID-19 cases and total infections (reported and non-reported) for each county and day for 30 days after the start date. This output is used to drive new SARS-CoV-2 infections for the ABM.

Acknowledgements: We are grateful for the support and input from Susan Eversole, Stacy Endres-Dighe, Georgiy Bobashev, and Alex Giarrocco of RTI International and from our UNC

Health collaborators, and our public health collaborators. This activity was based on a model originally developed through support from CDC's Modeling Infectious Disease in Healthcare (MInD-Healthcare) Network.**References**

1. Grimm V, Railsback S, Vincenot C, et al. The odd protocol for describing agent-based and other simulation models: A second update to improve clarity, replication, and structural realism. *Journal of Artificial Societies and Social Simulation*. 2020;23(2).

2. Slayton R, O'Hagan J, Barnes S, et al. Modeling infectious diseases in healthcare network (mind-healthcare) framework for describing and reporting multidrug-resistant organism and healthcare-associated infections agent-based modeling methods. *Clinical Infectious Diseases*. Published online 2020.

3. NC Department of Health and Human Services, Division of Health Services Regulation. Hospitals by county 2018. Published online 2018. https://info.ncdhhs.gov/dhsr/data/hllistco.pdf

4. Centers for Disease Control and Prevention. CDC WONDER mortality data. Published online 2017. https://wonder.cdc.gov/

5. Jones K, Munoz B, Rineer J, Bobashev G, Hilscher R, Rhea S. On calibrating a microsimulation of patient movement through a healthcare network. *In 2019 Winter Simulation Conference (WSC)*. Published online 2019:205-214.

6. Rhea S, Hilscher R, Rineer J, et al. Creation of a geospatially explicit, agent-based model of a regional healthcare network with application to *Clostridioides difficile* infection. *Health Security*. 2019;17(4).

## Appendix A: Parameters

*COVID-19 Parameters*

The parameter names are the same as the parameters specified in the parameters file used in model runs.

| Parameter | Description | Value | Source |
|---|---|---|---|
| r0 | Reproduction number; the average number of agents who will become infected with SARS-CoV-2 from one agent with SARS-CoV-2 infection | Varies | Sampled |
| min r0; max r0 | Minimum and maximum $R_0$ used as bounds for selecting $R_0$ | [1 - 1.2], [1.2 - 1.4], [1.4 - 1.6] | 23–28 |
| los mean | Mean length of stay (LOS): The average number of days that admitted COVID-19 agents spend in a hospital. Used in a truncated normal distribution for sampling agent LOS | 3 | 29 |
| los std | Standard deviation of LOS: The standard deviation in number of days that admitted COVID-19 agents spend in a hospital. Used in a truncated normal distribution for sampling agent LOS | 5 | 29 |
| los min | Minimum LOS: The minimum number of days that admitted COVID-19 agents spend in a hospital. Used in a truncated normal distribution for sampling agent LOS | 0 | 29 |
| los max | Maximum LOS: The maximum number of days that admitted | 50 | 29 |

| Parameter | Description | Value | Source |
|---|---|---|---|
| | COVID-19 agents spend in a hospital. Used in a truncated normal distribution for sampling agent LOS | | |
| infection duration | Length of infection in days, used for SEIR model and for calculating recovery day among COVID-19 agents not admitted to a hospital | 14 | Expert Input |
| ratio to hospital | Proportion of agents with asymptomatic, mild, or moderate symptoms that seek hospitalization | 0.0 | Expert Input |
| p tested | Proportion of infected agents that are tested | 0.1 | Inverse of initial case multiplier |
| icu with ventilator p | Probability that a critically ill agent in an ICU also needs a ventilator | 0.75 | [30] |
| initial case multiplier | Multiplier representing ratio of unreported infections to reported cases prior to start of the model | 10 | [16–19] |
| hospital to nh non icu | Probability of COVID-19 agent going to a nursing home after discharge from a non-ICU bed | 0.1 | Expert Input |
| hospital to nh icu | Probability of COVID-19 agent going to a nursing home after discharge from an ICU bed | 0.2 | Expert Input |
| prop of covid hospitalized to icu | Proportion of hospitalized COVID-19 agents in need of an ICU bed | 0.25 | [21], Expert Input |
| ages get covid | Distribution of positive COVID-19 reported cases by age | [Age 0 - 49: 0.396], [50-64: 0.328], [65+: 0.276] | Bayesian Calculation (described below) |
| p covidhospitalized tested concurrent | Probability of hospitalization by age given a positive, tested SARS-CoV-2 infection with comorbidities | [Age 0 - 49: 0.0], [Age 50 - 64: 0.4609], [Age 65+: 0.411] | Bayesian Calculation (described below) |
| p covidhospitalized tested notconcurrent | Probability of hospitalization by age given a positive, tested SARS-CoV-2 infection without comorbidities | [Age 0 - 49: 0.0367], [Age 50 - 64: 0.035], [Age 65+: 0.1213] | Bayesian Calculation (described below) |
| p covidhospitalized nottested concurrent | Probability of hospitalization by age given a positive, untested SARS-CoV-2 infection with comorbidities | [Age 0 - 49: 0.0], [Age 50 - 64: 0.0651], [Age 65+: 0.058] | Bayesian Calculation (described below) |
| p covidhospitalized nottested notconcurrent | Probability of hospitalization by age given a positive, untested SARS-CoV-2 infection without comorbidities | [Age 0 - 49: 0.0052], [Age 50 - 64: 0.0049], [Age 65+: 0.0171] | Bayesian Calculation (described below) |

*Bayesian Calculations*

Bayesian probabilities are used in the model to determine the probability of infection given age and the probability of hospitalization given testing status, age, and presence of comorbidities. Using a Bayesian approach allows for the conditional probabilities needed to account for the underlying variables of interest (testing status, age, presence of comorbidities). Model results can be disaggregated by the underlying variables that are used in the Bayesian analysis, with a more accurate interpretation than without the Bayesian approach, noting the limitations and assumptions of the Bayesian method.

Probability (Infection | Age)

The probability of infection given the age of the agent is important for creating a realistic distribution of agents among COVID-19 agents. This probability is used when selecting agents for infection (i.e., assignment to COVID-19 status). Each agent receives a probability of infection with SARS-CoV-2 given their age. The equation used to calculate this probability is:

*P(infection | age ) = (P(age | infection) * P(infection)) / P(age)*

The results of this equation are scaled so that the sum of P(infection | age) is 1, since the number of infections is specified by the SEIR equation. The effect of this scaling means that P(infection) does not impact the outcome.

**Equation Components**

- *P(age | infection)*: Calculated using public dashboard.[33]
- *P(infection)*: Calculated as the cumulative estimated infections for the model divided by the number of individuals in the synthetic population. This parameter changes for each scenario, although the scaling to 1 eliminates the impact of this parameter.
- *P(age)*: Distribution of age groups in the synthetic population.

Probability(Hospitalization | Age & Comorbidities) by Testing Status

The probability of hospitalization is related to age and presence of comorbidities, and that tested cases are more likely to seek hospitalization than unreported infections.[31] We use the same Bayesian equation to separately calculate the probability of hospitalization given age and comorbidities depending on testing status. Note that we do not currently account for any changes in testing status (i.e., an untested infection could receive a test after hospitalization).

The Bayes equation is:

*P(Hospitalization | Age & Concurrent Conditions) = P(Age & Concurrent Conditions | Hospitalization) * P(Hospitalization) / P(Age & Concurrent Conditions)*

**Equation Components**

- *P(Age & Concurrent Conditions | Hospitalization)*: Estimates of P(Age | Hospitalization) and P(Concurrent Conditions | Hospitalization) which were multiplied to estimate P(Age & Concurrent Conditions | Hospitalization). Note that we assume conditional independence of age and comorbidities given hospitalization for COVID-19 to make this calculation. This assumption may not hold if, for example, agents ≥ 65 years of age hospitalized with COVID-19 are more likely to have comorbidities than 50- to 64-year-olds hospitalized with COVID-19. Therefore, we do not suggest drawing robust conclusions from the values of the Bayesian probabilities until further data are obtained to confirm or refute conditional independence. The same values were used for both tested and untested infections.
- *P(Hospitalization)*: This component reflects the difference in likelihood of hospitalization between tested and untested infections. Based on expert input, the probability of hospitalization among tested infections is estimated at 8.5% and among untested infections at 1.2%.

- *P(Age & Concurrent Conditions)*: This component is calculated from the joint distribution of age and concurrent conditions in the synthetic population. Note the limitation that individuals <50 years of age in the synthetic population are not eligible to have comorbidities.